\documentclass[conference]{IEEEtran}
\IEEEoverridecommandlockouts

\usepackage{cite}
\usepackage{amsmath,amssymb,amsfonts}
\usepackage{algorithmic}
\usepackage{graphicx}
\usepackage{textcomp}
\usepackage{xcolor}
\usepackage{multirow}
\def\BibTeX{{\rm B\kern-.05em{\sc i\kern-.025em b}\kern-.08em
    T\kern-.1667em\lower.7ex\hbox{E}\kern-.125emX}}
\begin{document}

\title{Self-Supervised Point Cloud Completion based on Multi-View Augmentations of Single Partial Point Cloud
\thanks{$^*$Corresponding author.}
}

\author{
\IEEEauthorblockN{1\textsuperscript{st} Jingjing Lu}
\IEEEauthorblockA{\textit{Hunan University} \\
Changsha, China \\
lujingjing000@hnu.edu.cn} \\

\IEEEauthorblockN{4\textsuperscript{th} Zhuo Tang$^*$}
\IEEEauthorblockA{\textit{Hunan University} \\
Changsha, China \\
ztang@hnu.edu.cn}

\and

\IEEEauthorblockN{2\textsuperscript{nd} Huilong Pi$^*$}
\IEEEauthorblockA{\textit{Hunan University} \\
Changsha, China \\
phl880217@hnu.edu.cn} \\

\IEEEauthorblockN{5\textsuperscript{th} Ruihui Li}
\IEEEauthorblockA{\textit{Hunan University} \\
Changsha, China \\
liruihui@hnu.edu.cn}

\and

\IEEEauthorblockN{3\textsuperscript{rd} Yunchuan Qin}
\IEEEauthorblockA{\textit{Hunan University} \\
Changsha, China \\
qinyunchuan@hnu.edu.cn} \\
}

\maketitle

\begin{abstract}
Point cloud completion aims to reconstruct complete shapes from partial observations. Although current methods have achieved remarkable performance, they still have some limitations: Supervised methods heavily rely on ground truth, which limits their generalization to real-world datasets due to the synthetic-to-real domain gap. Unsupervised methods require complete point clouds to compose unpaired training data, and weakly-supervised methods need multi-view observations of the object. Existing self-supervised methods frequently produce unsatisfactory predictions due to the limited capabilities of their self-supervised signals. To overcome these challenges, we propose a novel self-supervised point cloud completion method. We design a set of novel self-supervised signals based on multi-view augmentations of the single partial point cloud. Additionally, to enhance the model's learning ability, we first incorporate Mamba into self-supervised point cloud completion task, encouraging the model to generate point clouds with better quality.
Experiments on synthetic and real-world datasets demonstrate that our method achieves state-of-the-art results. 
\end{abstract}

\section{Introduction}
Point clouds collected from real-world scenes often have missing regions due to sensor limitations or occlusions. 
Point cloud completion aims to recover these incomplete regions of partial shapes to reconstruct more complete shapes. It has great potential for applications in robotics and other fields.

In recent years, point cloud learning based on deep learning \cite{DGCNN} has greatly promoted the emergence of learning-based point cloud completion methods \cite{pcl2pcl}.

Supervised methods \cite{pcn} utilize paired training data consisting of partial point clouds and their ground truth to learn the mapping from incomplete shapes to complete shapes.
Unsupervised methods \cite{pcl2pcl,cGAN,Cycle4Completion} typically leverage partial point clouds along with unpaired, clean, complete example point clouds within the same category to train category-specific models.
Weakly-supervised method \cite{WeakPCN} employs unaligned partial observations from different viewpoints of the same object as weakly-supervised signals. 

These methods have certain drawbacks: (1) They rely on sufficient prior information during training for accurate shape reconstruction. Ground truth can offer supervised methods precise geometric information of the object. 
Complete example point clouds can serve as clean reference samples in the same category for unsupervised methods to generate predicted shapes.
Multi-view partial point clouds can provide weakly-supervised methods with detailed information of the object from different perspectives.
(2) The reliance on prior information limits the size of datasets and the diversity of data categories due to labor costs and equipment expenses, thereby constraining the model's scalability and applicability.

\begin{figure}[t]
\centering
    \includegraphics[width=7.5cm]{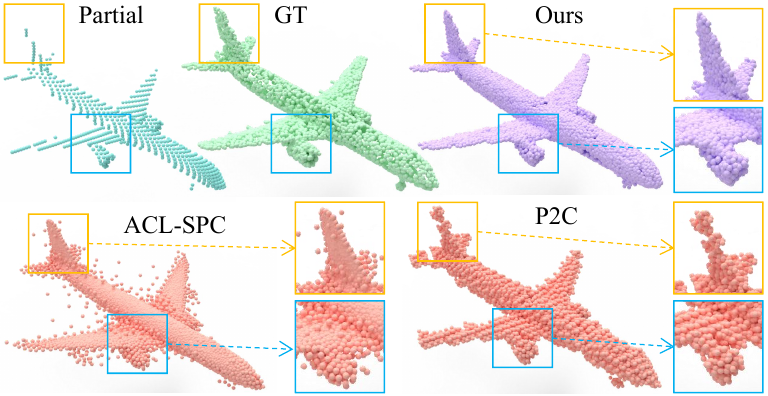}
    \vspace{-0.3cm}
    \caption{The predicted results of existing self-supervised methods and ours. }
    \vspace{-0.65cm}
    \label{fig1_compare}
\end{figure}

Self-supervised methods \cite{ACL-SPC,P2C} aim to tackle these challenges by utilizing only the single partial point cloud without any prior information for training. 
However, the predicted results of existing self-supervised methods are often unsatisfactory. As shown in Fig.~\ref{fig1_compare}, the predicted point clouds generated by ACL-SPC \cite{ACL-SPC} and P2C \cite{P2C} are blurry, lacking fine structural details. There are two main reasons for these poor results. (1) The self-supervised signals used by these methods have weak self-supervised capabilities. They primarily rely on the initial partial point cloud and the intermediate representations in training as their main self-supervised signals. Unfortunately, each object typically provides only one partial shape for training, leading to singular incompleteness that can cause the model to overfit to specific shapes, which hinders their generalization to other types of incomplete shapes. Additionally, the intermediate representations are always noisy, which can confuse the model.
(2) The model architectures used in these methods have limited capabilities. Previous methods adopt architectures designed for other tasks, which may not generalize well in self-supervised point cloud completion task. For example, ACL-SPC employs PolyNet \cite{polynet}, which is tailored for 3D shape recognition task, while P2C utilizes the encoder from PCN \cite{pcn}, which was developed for supervised point cloud completion task. 

\begin{figure*}
    \centering
    \includegraphics[width=0.9\linewidth]{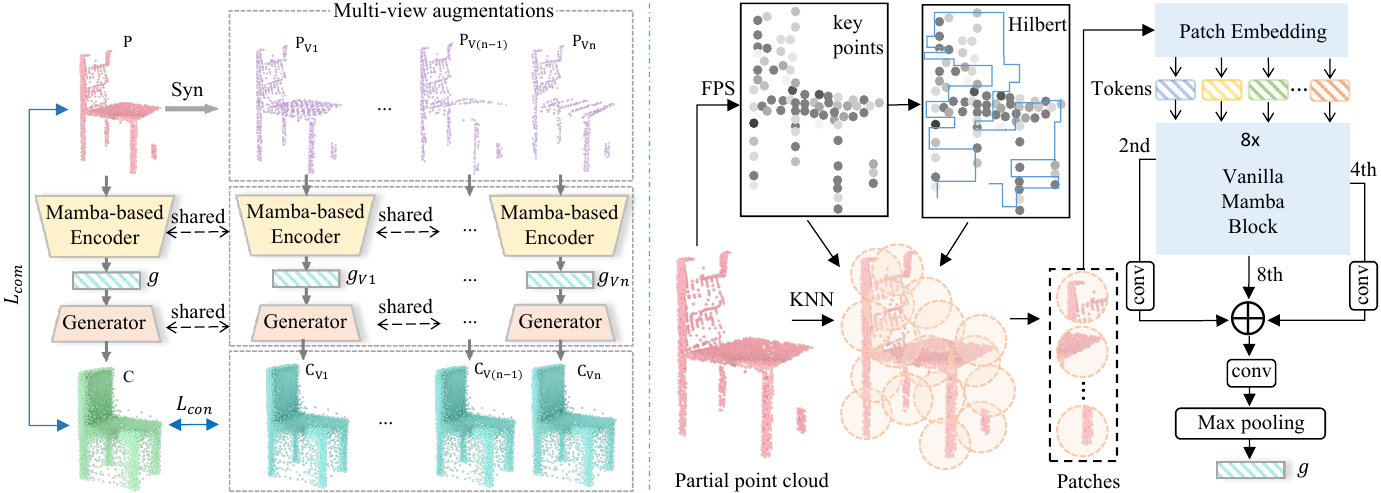}
    \vspace{-0.2cm}
    \caption{The architecture (left) and the Mamba-based encoder (right) of our method.} 
    \label{figure:architecture}
    \vspace{-0.5cm}
\end{figure*}

To overcome these challenges, we propose a novel self-supervised method for point cloud completion as shown in Fig.~\ref{figure:architecture}.
To improve the self-supervised capabilities of our method, we design a set of novel and effective self-supervised signals based on multi-view augmentations of the single partial point cloud, encouraging the model to generate point clouds with more accurate shapes. Specifically, we synthesize multiple partial shapes from different viewpoints of the initial partial point cloud $P$, creating multi-view augmented partial point clouds $P_{Vi} (\forall i\in[1,n])$ with clean and sparser points. By introducing diversity in incompleteness into the training data through the incorporation of various incomplete point clouds of the object, our model can learn to recognize and adapt to different forms of incompleteness. This approach significantly enhances its overall robustness and generalization capabilities across a range of incomplete shapes of the object.
Since both the initial and synthetic partial shapes are derived from the same object, their predicted point clouds are expected to ideally be identical. This correlation makes $C_{Vi}$ effective as self-supervised signals for training, encouraging the model to adaptively respond to variations in partial inputs by adjusting its parameters to generate the same outputs.
Additionally, we also use initial partial point cloud $P$ as a effective self-supervised signal. We can observe from Fig.~\ref{figure:architecture} that $P$ is relatively more complete compared to $P_{Vi}$, providing intrinsic geometric priors and structural details for shape completion in training. Using $P$ to supervise the completion of $P_{Vi}$ helps the model become robust to various types of incompleteness.

Recently, Mamba \cite{mamba} introduced the selective SSM mechanism and demonstrated great potential in point cloud analysis tasks by leveraging its global modeling capabilities. Inspired by \cite{pointmamba}, we first apply Mamba to the self-supervised point cloud completion task by designing a Mamba-based encoder to improve the encoder’s capability to extract global and local features, thereby further enhancing the model’s inference capabilities. We can observe from Fig.~\ref{fig1_compare} that the predicted shape of our method is more accurate and has finer-grained local details than other self-supervised methods.

We start by employing Farthest Point Sampling (FPS) to select the key points from the partial point cloud. Next, we apply Hilbert curves, a type of space-filling curve, to serialize these key points. Based on these serialized points, KNN is applied to partition the partial point cloud into local patches, which are fed into the patch embedding layer to generate serialized patch tokens. The primary module of the encoder is designed to be simple, consisting of eight basic, non-hierarchical Mamba blocks, which take these tokens as input and ultimately generate the global feature of the partial shape.

The main contributions of our method are as follows: 
\begin{itemize}
    \item We propose a new self-supervised method for point cloud completion leveraging only the single partial point cloud without any prior information.
    \item We introduce a set of novel and effective self-supervised signals based on multi-view augmentations of the single partial point cloud for training.
    \item We first apply Mamba to the self-supervised point cloud completion to enhance the model's inference capabilities.
    \item We conduct experiments on both synthetic and real-world datasets, demonstrating that our method outperforms previous methods, achieves state-of-the-art performance and generates point clouds with more accurate shapes.
\end{itemize}

\section{Related work}
\textbf{Unsupervised point cloud completion.} 
To eliminate the need for paired training data of supervised methods,
Pcl2Pcl \cite{pcl2pcl} leverages a generative adversarial network to train with unpaired data. Other methods \cite{Cycle4Completion,ShapeInversion} have been proposed to generate more accurate shapes. 

\textbf{Self-supervised point cloud completion.} 
Self-supervised methods \cite{ACL-SPC,P2C} are practical as they leverage effective self-supervised signals to generate a complete shape from a single partial input captured at an unknown viewpoint, without any prior information. However, the quality of the predicted point clouds generated by these methods is often unsatisfactory. To enhance the visual quality of the predictions, we design a set of novel self-supervised signals based on multi-view augmentations of the single partial point cloud, encouraging the model to generate point clouds with more accurate shapes.

\section{Method}
Fig.~\ref{figure:architecture} illustrates the architecture and training process of our method and the architecture of our Mamba-based encoder.

\subsection{Problem analysis and formulation}

To achieve self-supervised point cloud completion, we design a learnable model, denoted by $F_{\Theta}$, into which the partial point cloud $P\in\mathbb{R}^{N_P\times3}$ is fed as follows:
\begin{equation}\label{eq:eq1}
C=F_{\Theta}(P),
\end{equation}
where $C\in\mathbb{R}^{N_C\times3}$ represents the predicted point cloud, $N_P$ refers to the number of points in the input point cloud, and $N_C$ refers to the number of points in the output point cloud.

Then, we leverage a non-learnable partial shapes synthesizer $Syn$ to generate $P_{Vi}\in\mathbb{R}^{N_{Pvi}\times3}$ from $n = 8$ random viewpoints of the initial partial point cloud $P$. 
We randomly select viewpoints within the specified azimuth range ([0°, 360°]) and elevation range ([-20°, 40°]) using the uniform distribution.
The generated $P_{Vi}$ serve as multi-view augmented partial inputs for our model.
\begin{equation}\label{eq:eq2}
P_{Vi}=Syn(P, Vi), \forall i\in[1,n],
\end{equation}
where $Vi$ is a set of random viewpoints of $P$. 
$P_{Vi}$ are detached from computation graph and do not track operations for gradient computation during training. We feed $P_{Vi}$ into $F_{\Theta}$ to generate $n$ predicted point clouds $C_{Vi}\in\mathbb{R}^{N_C\times3}$.
\begin{equation}\label{eq:eq3}
C_{Vi}=F_{\Theta}(P_{Vi})=F_{\Theta}(Syn(P, Vi)), \forall i\in[1,n].
\end{equation}

\subsection{Generation of multi-view augmentations}
We use a non-learnable function $Syn$ to synthesize multi-view augmentations by projecting the initial partial point cloud $P$ onto depth maps from eight random viewpoints. We then back-project these depth maps into 3D space. Specifically, (1) Depth map generation: Choose a random viewpoint from which to "observe" $P$. This viewpoint acts like a camera position in 3D space. From the selected viewpoint, project $P$ onto a 2D plane. Each point in $P$ is mapped to a 2D coordinate on this plane, with the third dimension (depth) representing the distance from the viewpoint to the surface of $P$. The resulting depth map captures how $P$ appears from that specific viewpoint, often resembling a grayscale image where pixel values correspond to depth information. (2) Back-projection into 3D: Convert the 2D depth map of $P$ back into a 3D point cloud by reversing the projection process. Each pixel of the depth map is transformed back to its corresponding 3D coordinates, creating a new point cloud. This new point cloud is a partial shape of the initial partial point cloud $P$ because it only includes points visible from the chosen viewpoint, simulating occlusion effects and missing data that occur in real-world scenarios. The synthetic multi-view augmentations of $P$ can be found in supplementary material (SM) sec. S5.

\begin{table*}
    \centering
    \caption{Quantitative comparisons on the synthetic ShapeNet dataset. All the values are multiplied by 100, lower is better.} 
    \vspace{-0.3cm}
    \label{table:shapenet}
    \scalebox{1.0}{
    \begin{tabular}{c c ccc c ccc c ccc c ccc}
	\cline{1-17}
	\multirow{2}{*}{\textbf{Supervision}} & \multirow{2}{*}{\textbf{Methods}} & \multicolumn{3}{c}{\textbf{Airplane}} && \multicolumn{3}{c}{\textbf{Car}} && \multicolumn{3}{c}{\textbf{Chair}} && \multicolumn{3}{c}{\textbf{Average}}   \\
        && 
        P$\downarrow$& C$\downarrow$ & CD$\downarrow$ && P$\downarrow$ & C$\downarrow$ & CD$\downarrow$ && P$\downarrow$ & C$\downarrow$ & CD$\downarrow$ && P$\downarrow$ & C$\downarrow$ & CD$\downarrow$ \\
	\cline{1-17}
        \multirow{3}{*}{Unsupervised} & DPC\cite{DPC}
            & {-} & {-} & {3.91} &
            & {-} & {-} & {3.47} &
            & {-} & {-} & {4.30} &
            & {-} & {-} & {3.89} \\
        
        & Gu et al.\cite{WeakPCN}  
            & {\textbf{0.91}} & {1.05} & {1.95} &
            & {1.27} & {1.41} & {2.68} &
            & {1.69} & {1.64} & {3.33} &
            & {1.29} & {1.36} & {2.65} \\
        
        & PointPnCNet\cite{PointPnCNet}
            & {1.58} & {1.74} & {3.32} &
            & {1.98} & {2.98} & {4.96} &
            & {2.72} & {2.68} & {5.40} &
            & {1.75} & {2.46} & {4.56} \\
        & Opt-DE\cite{Optde}
            & {1.12} & {0.95} & {2.07} &
            & {1.46} & {1.52} & {2.98} &
            & {2.56} & {1.68} & {4.24} &
            & {2.12} & {1.56} & {3.68} \\
        \cline{1-17}
        \multirow{3}{*}{Self-supervised}  & ACL-SPC\cite{ACL-SPC} 
            & {1.20} & {\textbf{0.80}} & {2.01} & 
            & {1.65} & {1.28} & {2.93} &
            & {2.25} & {1.46} & {3.71} &
            & {1.70} & {1.18} & {2.88} \\
	& P2C\cite{P2C}    
            & 1.70  & 0.90  & 2.60  &
            & 1.80  & 2.00  & 3.80  & 
            & 3.33  & 1.50  & 4.83  & 
            & 2.28  & 1.47  & 3.74  \\
	& \textbf{Ours} 
            & 1.03 & 0.88 & \textbf{1.91} &
            & \textbf{0.83} & \textbf{0.52} & \textbf{1.35} & 
            & \textbf{1.05} & \textbf{0.96} & \textbf{2.01} & 
            & \textbf{0.97} & \textbf{0.79} & \textbf{1.76} \\
	\cline{1-17}
    \end{tabular}}
    \label{table}
    \vspace{-0.5cm}
\end{table*}

\subsection{Mamba-based Encoder and Generator}
As illustrated in Fig.~\ref{figure:architecture}, we first utilize FPS to select $m$ key points from the partial point cloud. FPS can select points that are maximally spaced apart, ensuring a comprehensive representation of the point cloud’s structure and capturing the essential geometric characteristics of the point cloud.

Typically, the arrangement of sampled key points is arbitrary without any particular sequence. This lack of order tends to be inconsequential for earlier Transformer-based models, as they possess an order-invariant property through their self-attention mechanisms. 
However, for selective state space models, i.e., Mamba, the nature of unidirectional modeling makes it challenging to effectively analyze unstructured and disordered point clouds. 
Therefore, we implement a serialization process to convert the unordered key points into a sequential format. 
We choose to employ a space-filling curve, specifically the Hilbert curve \cite{Hilbert}, for this serialization. The Hilbert curve is particularly effective because it maintains spatial locality, ensuring that points close to one another in 3D space remain proximate in the serialized sequence. This preserves the inherent geometric relationships within the key points, facilitating accurate feature representation of point clouds.

After serialization, we apply the K-Nearest Neighbors (KNN) algorithm to form patches around each key point by selecting the $k$ nearest points in the neighborhood of each key point. To aggregate local information, points within each patch are normalized by subtracting the key point to obtain relative coordinates. 
These patches are processed by the patch embedding layer implemented using a lightweight DGCNN\cite{DGCNN} to generate a contextually rich feature representation for each patch, obtaining serialized patch tokens.

The primary module of our encoder is designed to be extremely straightforward, consisting of eight plain and non-hierarchical Mamba blocks. 
We use Mamba because it offers linear complexity and is well-suited for sequential modeling tasks due to its efficient handling of sequences with the selective state space models. Mamba allows us to retain the benefits of global contextualization present in transformer architectures but without the high computational cost associated with attention mechanisms. This makes it particularly useful for point cloud completion, where efficiency and performance are both critical. By employing Mamba, we ensure that the system remains efficient in processing while maintaining powerful feature extraction capabilities.
Mamba's architecture permits seamless integration into the processing pipeline, allowing it to capture global dependencies across the serialized sequences. 
More details about the Mamba block are in SM sec. S2.

To fully utilize the features from the middle layers, we extract features from the 2nd and 4th Mamba blocks, apply the $conv$ function to each, and concatenate the results with the 8th Mamba block's features. We then use the $conv$ function to adjust the channel dimensions. After a max pooling operation, we finally obtain a global feature $g$ with a dimension of 512.

We employ a generator comprising three fully connected layers with respective sizes of 1024, 1024, and \( N_C \times 3 \). The ReLU non-linear activation function is applied to the outputs of the first and second fully connected layers.

\subsection{Loss functions}
Our method incorporates two sets of effective self-supervised signals. First, we use initial partial shapes as self-supervised signals and employ weighted Chamfer distance loss \cite{PointPnCNet} $L_{com}$ between $P$ and $C$, where $||.||_2$ represents the $L_2$ norm, $x$ is the point of $C$ and $y$ is the point of $P$.
\begin{equation}\label{eq:Lcom}
\mathcal{L}_{com} = \frac{\alpha}{N_{C}}\sum_{x\in C} \min_{y\in P}||x-y||_2 + \frac{\beta}{{N_P}}\sum_{y\in P} \min_{x\in C}||y-x||_2
\end{equation}
We further utilize $C_{Vi}$ as effective self-supervised signals and calculate the consistency loss $L_{con}$ between $C_{Vi}$ and $C$.
\begin{equation}\label{eq:Lcons}
\mathcal{L}_{con} = \frac{1}{N_{C}\times n}\sum_{i=1}^{n}||C_{Vi}-C||^2_2
\end{equation}
The total loss $L$ is the weighted summation of $L_{com}$ and $L_{con}$. We set \( \gamma = 15 \), \(\alpha = 0.1 \), and \(\beta = 0.9\) during training.
\begin{equation}\label{eq:L}
\mathcal{L} = \mathcal{L}_{com} + \gamma\mathcal{L}_{con}
\end{equation}

\section{Experiments}
We conduct category-specific experiments on both synthetic and real-world datasets to demonstrate the effectiveness of our method. 
For a fair comparison, we implement the source codes and the same configurations of the compared methods, except for Gu et al. \cite{WeakPCN}, for which the source code is not available, so we cite their reported results.
For different datasets, the resolution of the partial point clouds may differ, but the resolution of the predicted point clouds is the same at 8,192 points.
More experimental results, training and inference details, the details of evaluation metrics, complexity and efficiency analysis, and robustness evaluation can be found in SM.

\subsection{Comparisons on synthetic dataset}

\textbf{Dataset and evaluation metrics. }
We conduct experiments on the ShapeNet \cite{ShapeNet} dataset, which is the largest and most widely used dataset in point cloud completion research. It comprises large-scale 3D point clouds depicted by CAD models spanning 55 categories. We focus on three categories: chair, airplane, and car, following \cite{WeakPCN, PointPnCNet, DPC, ACL-SPC}. 
The initial partial point clouds, and multi-view augmented partial point clouds contain 3,096 and 2,048 points, respectively.

Following \cite{WeakPCN, PointPnCNet, DPC, ACL-SPC}, we leverage three evaluation metrics: (1) Chamfer Distance (CD) measures the distance between the predicted shape and the ground truth. (2) Precision describes how well the predicted points are distributed compared to the ground truth. (3) Coverage indicates how much the missing parts of the partial point cloud are filled.

\textbf{Quantitative and qualitative comparisons. }
For the quantitative comparisons shown in Table~\ref{table:shapenet}, our method surpasses both unsupervised methods and self-supervised methods in the car and chair categories, as well as in average performance across all metrics, while achieving the best CD performance in the airplane category. 
Specifically, compared to the best performances of other methods, our method improves average performance by approximately 24.8$\%$ in precision, 33.1$\%$ in coverage, and 33.6$\%$ in CD, demonstrating the significant advancement made by our method.

\begin{figure}[t]
    \centering
    \includegraphics[width=1.0\linewidth]{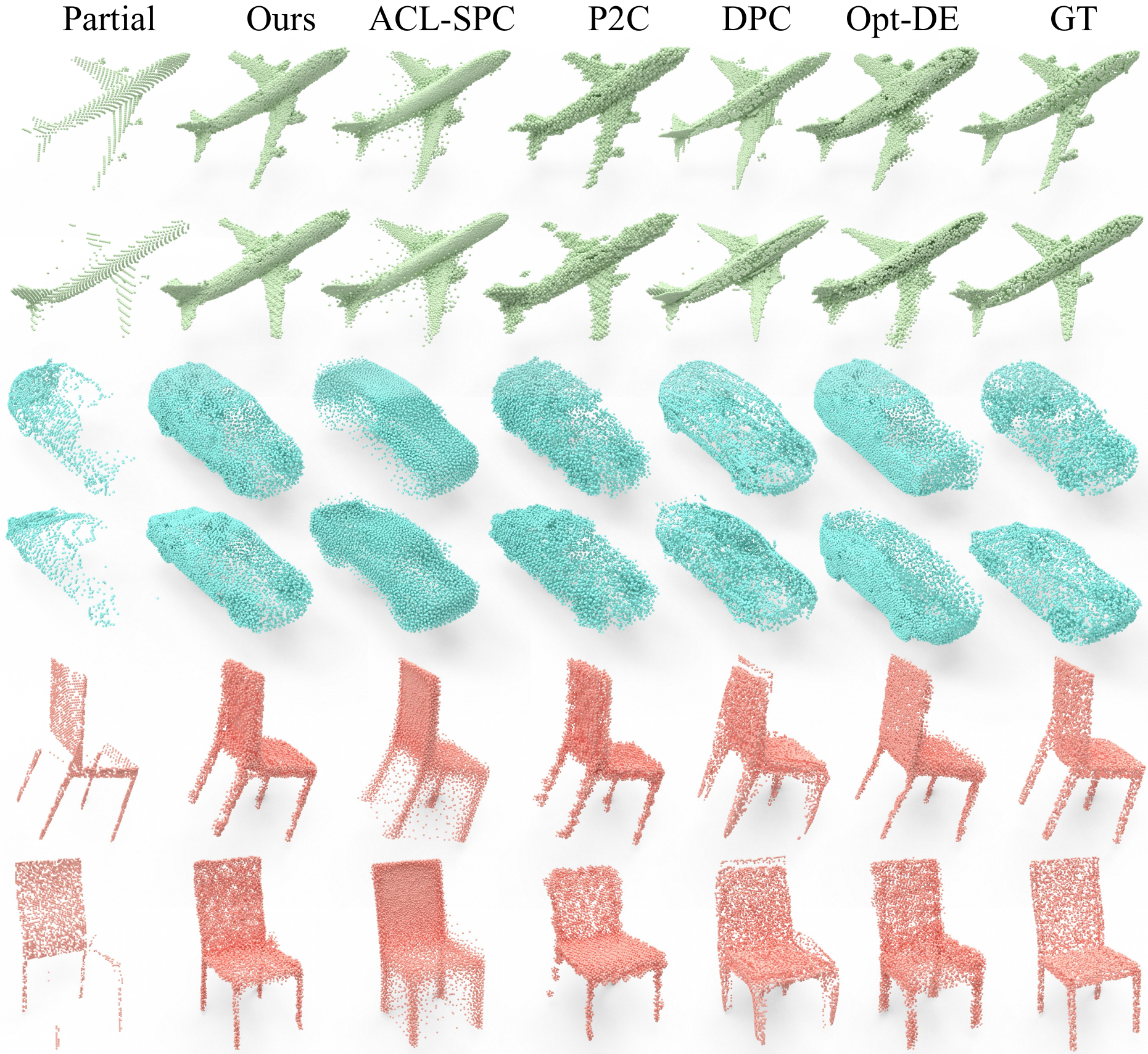}
    \vspace{-0.8cm}
    \caption{Qualitative comparisons on the ShapeNet dataset.}
    \label{figure:shapenet}
    \vspace{-0.5cm}
\end{figure}

For the qualitative comparisons visualized in Fig.~\ref{figure:shapenet}, our method effectively generates missing parts with fine-grained structures. The predicted shapes of our method are more reasonable and accurate than those of other methods. 
However, compared to the ground truth (GT), our predicted shapes are not sufficiently smooth and lack sharpness in their edges, particularly in the predicted missing parts. This issue arises from the limited self-supervised capabilities of our method. 

\begin{table*}
    \centering
    \caption{Quantitative comparisons on ScanNet, MatterPort3D and KITTI datasets. For UCD$\times$10000 and UHD$\times$100 , lower is better.} 
    \vspace{-0.3cm}
    \label{table:scanmatterkitti}
    \scalebox{1.0}{
    \begin{tabular}{c c cccc c cccc c cc}
	\cline{1-14}
	\multirow{3}{*}{\textbf{Supervision}} & \multirow{3}{*}{\textbf{Methods}} & \multicolumn{4}{c}{\textbf{ScanNet}} && \multicolumn{4}{c}{\textbf{MatterPort3D}} && \multicolumn{2}{c}{\textbf{KITTI}}  \\
        & & \multicolumn{2}{c}{\textbf{Chair}} & \multicolumn{2}{c}{\textbf{Table}} && \multicolumn{2}{c}{\textbf{Chair}} & \multicolumn{2}{c}{\textbf{Table}} && \multicolumn{2}{c}{\textbf{Car}}  \\
        
        && {UCD$\downarrow$} & {UHD$\downarrow$} & {UCD$\downarrow$} & {UHD$\downarrow$} && {UCD$\downarrow$} & {UHD$\downarrow$} & {UCD$\downarrow$} & {UHD$\downarrow$} && {UCD$\downarrow$} & {UHD$\downarrow$} \\
	\cline{1-14}
            \multirow{6}{*}{Unsupervised} 
            & pcl2pcl\cite{pcl2pcl} 
                & 17.3 & 10.1 & 9.1 & 11.8 && 15.9 & 10.5 & 6.0 &  11.8 && 9.2 &  14.1 \\
            & ShapeInversion\cite{ShapeInversion} 
                & 3.2 & 10.1 & 3.3 & 11.9 && 3.6 & 10.0 & 3.1 & 11.8 && 2.9 & 13.8 \\
            & $+$UHD\cite{ShapeInversion}
                & 4.0 &9.3 & 6.6 & 11.0 && 4.5 & 9.5 & 5.7 & 10.7 && 5.3 & 12.5 \\
            & Cycle4Comp.\cite{Cycle4Completion} 
                & 5.1 & 6.4 & 3.6 & 5.9	&& 8.0 & 8.4 & 4.2 & 6.8 && 3.3 & 5.8 \\
            & DE\cite{Optde}
                & 2.8 & 5.4 & 2.5 & 5.2 && 3.8 & 6.1 & 2.5 & 5.4 && 1.8 & 3.5 \\
            & OptDE\cite{Optde} 
                & 2.6 & 5.5 & 1.9 & 4.6 && 3.0 & 5.5 & 1.9 & 5.3 && 1.6 & 3.5 \\

        \cline{1-14}
        \multirow{3}{*}{Self-supervised}  
        & ACL-SPC\cite{ACL-SPC} & 1.4 &  4.7 &  1.8 & 5.1 && 1.8 & 4.8 & 2.1 & 4.9 && 2.0 & 4.9  \\
        & P2C\cite{P2C}         & 2.1 &  4.5 &  1.9 & 4.7 && 1.2 & 3.9 & 1.9 & 4.5 && 1.9 & 4.8   \\
	& \textbf{Ours} & \textbf{0.9} & \textbf{3.7} & \textbf{0.5} & \textbf{3.1} &&
                \textbf{0.6} & \textbf{3.4} & \textbf{0.6} & \textbf{3.5} && 
                \textbf{0.9} & \textbf{3.5}  \\
	\cline{1-14}
    \end{tabular}}
    \vspace{-0.5cm}
\end{table*}

\subsection{Comparisons on real-world datasets}
\textbf{Datasets and evaluation metrics. }

To evaluate the scalability and robustness of our method, we conduct experiments on real-world datasets that present diverse and challenging scenarios. These datasets consist of point clouds collected from actual scenes using radar or depth cameras, reflecting the incompleteness and noise in point clouds caused by factors such as occlusion and lighting in the environment.
We conduct experiments on three real-world datasets: ScanNet \cite{ScanNet}, MatterPort3D \cite{Matterport3D}, and KITTI \cite{geiger2012we}. ScanNet and MatterPort3D datasets offer 3D reconstructions of indoor scenes, and we focus on chairs and tables, following \cite{pcl2pcl,Optde,ShapeInversion,ACL-SPC}. The KITTI dataset showcases outdoor scenes primarily featuring cars. 
The initial partial point clouds and multi-view augmented partial point clouds contain 2,048 and 1,024 points, respectively.
We use Unidirectional Chamfer Distance (UCD) \cite{Cycle4Completion} and Unidirectional Hausdorff Distance (UHD) \cite{UHD} as the evaluation metrics, following \cite{pcl2pcl, ShapeInversion, Optde,ACL-SPC}.
We also evaluate our method on the SemanticKITTI \cite{SemanticKITTI} dataset. 
The initial partial point clouds and multi-view augmented partial point clouds contain 1,024 and 800 points, respectively.
We also use precision, coverage, and CD mentioned above as the evaluation metrics.

\textbf{Quantitative and qualitative comparisons. }
For the quantitative comparisons in Table~\ref{table:scanmatterkitti}, our method shows superior performance across all three datasets, outperforming self-supervised and unsupervised methods by a significant margin in terms of UCD and UHD metrics. Specifically, 6 out of 10 results improve by more than 32$\%$ compared to the best 
results obtained by other methods. Particularly, in the UCD metric, it achieves approximately 72$\%$ improvement for the ScanNet table, 50$\%$ for the MatterPort3D chair, 68.4$\%$ for the MatterPort3D table, and 43.8$\%$ for the KITTI car.
The results demonstrate that our method has made significant progress.

\begin{table}[t]
    \centering
    \caption{Quantitative comparisons on the SemanticKITTI dataset.} 
    \vspace{-0.3cm}
    \label{table:semanticKITTI}
    \scalebox{1.0}{
    \begin{tabular}{c c ccc }
	\cline{1-5}
	\textbf{Supervision} & \textbf{Method} & \textbf{P}$\downarrow$ & \textbf{C}$\downarrow$ & \textbf{CD}$\downarrow$ \\
	\cline{1-5}
        \multirow{3}{*}{Supervised} 
        & GRNet\cite{GRNET} 
            & \textbf{4.63}  & 6.90  & 11.53 \\
        & SFNet\cite{SnowflakeNet} 
            & 14.12 & 12.64 & 26.76 \\
        & pcn\cite{pcn}   
            & 9.83 & 17.96 & 27.79 \\
        \cline{1-5}
        \multirow{2}{*}{Unsupervised}  
        & Gu et al.\cite{WeakPCN}   
            &  8.70  & 10.70  & 19.40 \\
        & PointPnCNet\cite{PointPnCNet}   
            & 9.00 & 10.00 & 19.00 \\
        \cline{1-5}
        \multirow{3}{*}{Self-supervised}  
        & ACL-SPC\cite{ACL-SPC}  
            & 11.67 & 5.63 & 17.30 \\
        & P2C\cite{P2C}  
            & 12.33 & 6.23 & 18.56 \\
	& \textbf{Ours}   
            & 7.25 & \textbf{2.35} & \textbf{9.6}  \\
	\cline{1-5}
    \end{tabular}}
    \vspace{-0.5cm}
\end{table}

\begin{figure}[t]
    \centering
    \includegraphics[width=1.0\linewidth]{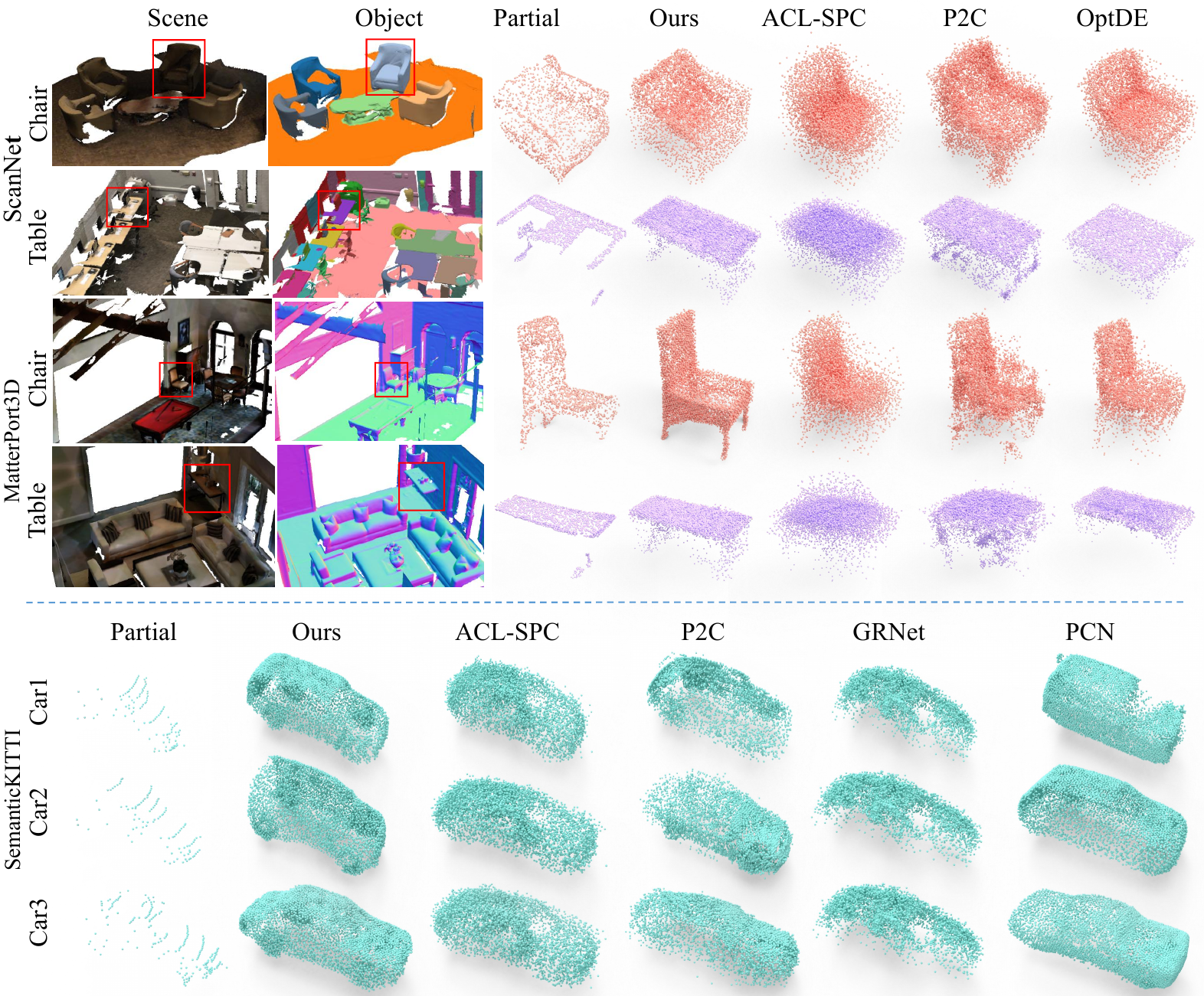}
    \vspace{-0.7cm}
    \caption{Visual results on ScanNet, MatterPort3D, and SemanticKITTI datasets.}
    \label{figure:realworld_3datasets}
    \vspace{-0.3cm}
\end{figure}

For the qualitative comparisons in Fig.~\ref{figure:realworld_3datasets}, our method predicts shapes with better visual quality than other methods.

However, the predicted shapes on real-world datasets often contain noisy points. As point clouds from real-world datasets are collected from actual scenes, they frequently introduce noise due to factors such as occlusion. Such noise poses a significant challenge when the model learns from these noisy partial inputs. As a result, even with effective training, the outputs may still exhibit the inherent noise present in the training data.
Despite this, the predicted results demonstrate the superior scalability and robustness of our method in real-world scenarios compared to other methods. Visual comparisons on the KITTI dataset can be found in SM sec. S9.

Additionally, we evaluate our method on the SemanticKITTI dataset. 
For the quantitative comparisons in Table~\ref{table:semanticKITTI}, our method outperforms supervised, unsupervised, and self-supervised methods in both the coverage and CD metrics. 
Specifically, our method achieves an improvement of about 58.3$\%$ in the coverage and an over 16.7$\%$ improvement in the CD compared to the best results from other methods.

For the qualitative comparisons shown in Fig.~\ref{figure:realworld_3datasets}, our method performs quite well on the SemanticKITTI dataset. Despite the partial point clouds being very sparse, making it challenging for humans to recognize them as parts of the car, our method effectively recovers the missing regions, generating complete car shapes.
In contrast, the predicted shapes of ACL-SPC lack diversity and local details. P2C struggles to fully recover missing regions, while the pre-trained models of supervised methods exhibit poor generalization, primarily due to the domain gap between synthetic and real-world datasets.

\subsection{Ablation study}
To evaluate the effect of each loss for training, we remove one loss at a time and conduct individual experiments. The testing results in Table~\ref{table:loss_effect} show that removing \( L_{con} \) significantly worsens coverage and CD metrics, which emphasizes its crucial role in training and further demonstrates the effectiveness of the novel self-supervised signals we introduce.

To evaluate the effectiveness of our Mamba-based encoder, we integrate different encoders into our method while keeping the decoder unchanged. Specifically, we substitute our Mamba-based encoder with the encoders from PCN, PolyNet, and Point-MAE, conducting experiments for each setup. As shown in Table~\ref{table:ablation_encoder}, all these ablation settings result in reduced performance compared to the full model. The experimental results demonstrate that our Mamba-based encoder improves the encoder's ability to extract both global and local features by effectively leveraging its efficient global modeling capacity, thereby enhancing the model's inference capabilities.

\begin{table}[t]
    \centering
    \caption{The effects of loss functions on ShapeNet dataset.} 
    \vspace{-0.3cm}
    \label{table:loss_effect}
    \scalebox{0.82}{
    \begin{tabular}{c | ccc | ccc | ccc}
	\cline{1-10}
	\multirow{2}{*}{\textbf{Loss}} & \multicolumn{3}{c|}{\textbf{Airplane}} & \multicolumn{3}{c|}{\textbf{Car}} & \multicolumn{3}{c}{\textbf{Chair}} \\
        & P$\downarrow$& C$\downarrow$ & CD$\downarrow$ & P$\downarrow$ & C$\downarrow$ & CD$\downarrow$ & P$\downarrow$ & C$\downarrow$ & CD$\downarrow$ \\
	\cline{1-10}
        $-L_{con}$ 
        & {1.32} & {19.12} & {20.44} & {3.66} & {20.18} & {23.84} & {3.72} & {21.58} & {25.30} \\
        $-L_{com}$  
        & {1.33} & {1.56} & {2.89} & {1.45} & {2.04} & {3.49} & {2.11} & {1.53} & {3.64} \\
	  $L$ 
        & {\textbf{1.03}} & {\textbf{0.88}} & \textbf{1.91} & {\textbf{0.83}} & \textbf{0.52} & {\textbf{1.35}} & {\textbf{1.05}} & \textbf{0.96} & \textbf{2.01} \\
         
	\cline{1-10}
    \end{tabular}}
    \vspace{-0.5cm}
\end{table}

\begin{table}[t]
    \centering
    \caption{Ablation study of Mamba-based encoder on the ShapeNet dataset. For CD \(\times\)100, lower is better.}
    \vspace{-0.3cm}
    \scalebox{0.8}{
		\begin{tabular}{c|cccc}
		  \hline
		  \textbf{Encoder} & \textbf{Airplane} & \textbf{Car} & \textbf{Chair} & \textbf{Average}  \\
		  \hline
            PCN'encoder\cite{pcn}   & 3.35 &  2.39 & 3.36 &  3.41 \\
		  PolyNet'encoder\cite{polynet}& 3.09 & 1.74 & 3.09 &  2.91 \\
            Point-MAE'encoder\cite{pointmae} & 2.01 & 1.51 & 2.22 &  1.89 \\
            \hline
            \textbf{Ours} & \textbf{1.91} & \textbf{1.35} & \textbf{2.01} & \textbf{1.76}  \\
            \hline
		\end{tabular}}
        \vspace{-0.4cm}
    \label{table:ablation_encoder}
\end{table}

\section{Conclusion}
We propose a novel and effective self-supervised method for point cloud completion by designing a set of novel self-supervised signals based on multi-view augmentations of the single partial point cloud. Experiments demonstrate that our method achieves state-of-the-art results.

\section*{Acknowledgment}
This work is supported by National Natural Science Foundation of China (Grant No.62202151) and National Natural Science Foundation of China (Grant No.62225205).

\bibliographystyle{IEEEbib}
\bibliography{reference}

\begin{center}
    {\small
    \textbf{© 2025 IEEE.} Personal use of this material is permitted. 
    Permission from IEEE must be obtained for all other uses, in any current 
    or future media, including reprinting/republishing this material for 
    advertising or promotional purposes, creating new collective works, 
    for resale or redistribution to servers or lists, or reuse of any 
    copyrighted component of this work in other works. \\
    
    This is the author's version of the work accepted to IEEE International 
    Conference on Multimedia and Expo (ICME) 2025. 
    The final version is to be published in \textit{Proceedings of IEEE ICME 2025} 
    and will be available via IEEE Xplore. \\
    }

\end{center}

\end{document}